# ISIC 2017 – Skin Lesion Analysis Towards Melanoma Detection

Matt Berseth
NLP LOGIX, LLC, matt.berseth@nlplogix.com

### INTRODUCTION

Our system addresses Part 1, Lesion Segmentation and Part 3, Lesion Classification of the ISIC 2017 challenge. Both algorithms make use of deep convolutional networks to achieve the challenge objective.

### LESION SEGMENTATION

**Preprocessing**
To prepare the images for the network, each of the training images was resized to 192 pixels by 192 pixels. To create additional training images, each of the training images was elastically distorted. For each of the original training images, four randomly generated elastic distorted images were generated and then resized down to 192 by 192 pixels. In addition, each training image was also rotated 90 degrees and additional elastic distortions were applied to the rotated images.

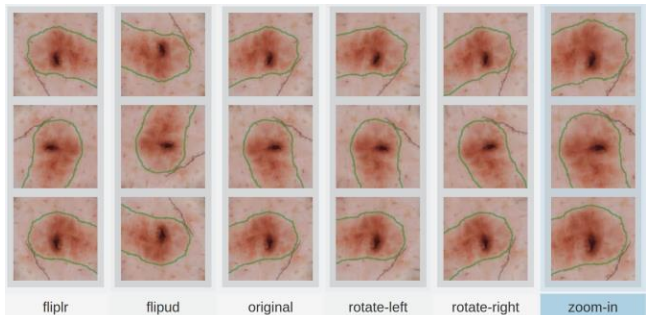
**Figure 1 Image and mask transformations**

After adding these additional distortions, an additional 9 training examples were created for each original training image. Each of the training masks was rotated and transformed in the same manner creating an additional 18,000 training images bringing the total number of training images to 20,000.

**Model Architecture**
A U-Net[1] architecture was used to provide a probability estimate for each pixel in the original image. The U-Net operated on an input image of 192 by 192 pixels and produced a probability map of the same dimensions. The U-Net included three down-sampling layers, a fully connected layer at the bottom of the `U` followed by three up-sampling layers. Unlike the network architecture in the original U-Net paper, the network used here produces an output map of the same dimensions as the input image.

Rectified linear units were used for all non-linearity's. Dropout was used at the bottom of the `U`.

**Training**
The network was trained using the Adam[2] optimization algorithm. The learning rate was initialized to 1e-4 and was not changed throughout the training process. Minibatches of size 20 were used and a custom weight map was provided with each minibatch that equalized the weights between the positive and negative classes.

10-fold cross validation was used to train the model. Each fold was selected randomly without stratification. Each of the original images, along with all distorted images were assigned to a single fold to ensure there wasn't any leaking between the train and test folds.

As the minibatches were selected, an additional set of transformations were applied to the mask and image at runtime. These transformations included the following:
- Flipping
- Rotation
- Zoom

Figure 1 shows the image distortions and transformations.

Training continued for a total of 200 epochs, after each epoch the Jaccard Index[3] was measured on the validation set and recorded. After the 200 epochs completed, the model selected for each fold corresponded to the iteration that maximized the Jaccard Index as computed on the validation data set.

**Postprocessing**
To score the validation and test submission sets, the best model from each of the 10 training folds was used. The full probability map from each of the models was averaged to generate the final probability map. A conditional random field was used to finetune the probability map, but this was ultimately discarded as it did not add significant performance.

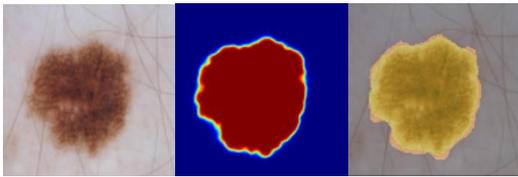

**Figure 2 left: Image ISIC_0000019.jpg, middle: probability map, right: estimated and actual segmentation**

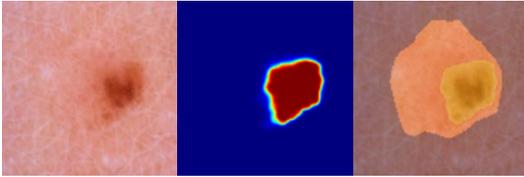

**Figure 3 left: Image ISIC_0000096.jpg, middle: probability map, right: estimated and actual segmentation**

### Results

Table 1 shows the best Jaccard Index score for each of the validation folds.

| Fold | Jaccard Index |
|------|---------------|
| 0 | 0.84017 |
| 1 | 0.82823 |
| 2 | 0.83541 |
| 3 | 0.83098 |
| 4 | 0.83721 |
| 5 | 0.82378 |
| 6 | 0.81489 |
| 7 | 0.84379 |
| 8 | 0.83180 |
| 9 | 0.84084 |
|   | **0.83271** |

**Table 1 Cross validation average Jaccard Index by fold**

## LESION CLASSIFICATION

### Preprocessing

To prepare the images for the network, each of the training images was cropped so the larger dimensions was the same size as the smaller dimension, and then resized to 256 pixels by 256 pixels. To create additional training images, each of the original images was rotated 90 and 270 degrees before resizing, bringing the total number of training images used to 6,000.

In addition, while analyzing the training images, two potential sources of data leaks were discovered:
- Images with colored gauze was visible were all the same class. See Figure 5.
- Images with a bright light on both left and right edges were all the same class. See Figure 7.

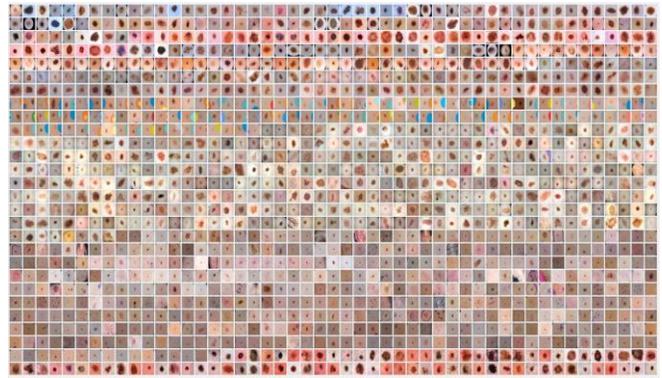

**Figure 4 All Nevus training images**

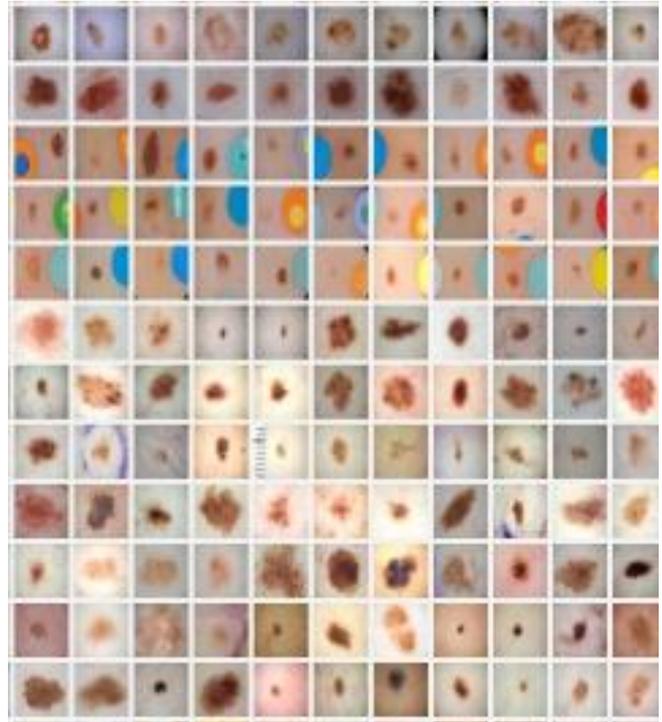

**Figure 5 Nevus training images with colored gauze**

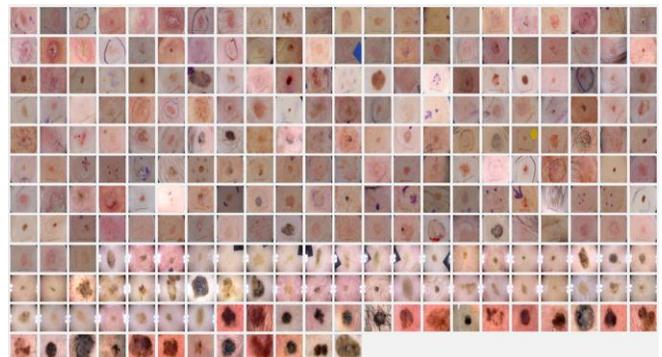

**Figure 6 All Seborrheic Keratosis training images**

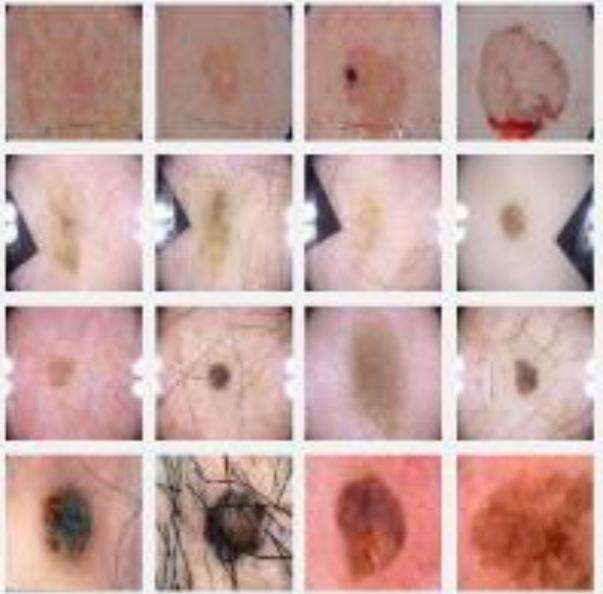
**Figure 7 Seborrheic Keratosis training images with bright light on edges**

To ensure the model was learning features about the lesion and not about the background artifacts of the image, each of these images from the training set was manually cropped to remove these objects.

### Model Architecture
A AlexNet[3] deep convolutional network architecture was used to classify the images. The network operated on an input image of 224 by 224 pixels and produced a probability distribution over the labels. Each of the training images was randomly cropped to 224 by 224 before presenting to the model. Unlike the traditional AlexNet architecture, the fully-connected layers contained 1024 neurons instead of the usual 4096. Dropout was used on the fully connected layers and rectified linear units were used for all non-linearity's.

### Training
The network was trained using the Adam[2] optimization algorithm. The learning rate was initialized to 1e-5 and was not changed throughout the training process. Minibatches of size 64 were used, each class represented an equal number of times.

10-fold cross validation was used to train the model. Each fold was selected randomly, but stratified across each of the lesion classes. Each of the original images, along with all the rotated images were assigned to a single fold to ensure there wasn't any leaking between the train and test folds.

As the minibatches were selected, an additional set of transformations were applied to the mask and image at runtime. These transformations included the following:
- Flipping
- Rotation
- Zoom

Training continued for a total of 300 epochs, after each epoch the AUC was measured on the validation set and recorded. After the 300 epochs completed, the model selected for each fold corresponded to the iteration that maximized the AUC as computed on the validation data set.

### Postprocessing
To score the validation and test submission sets, the best model from each of the 10 training folds was used. The probability rom each of the models was averaged to generate the final probability for each image. These probabilities were then joined to the patient demographic information and this was presented to a random forest model that generated the final probabilities.

### Results
Table 2 shows the best Jaccard Index score for each of the validation folds. These results are just based on the visual characteristics of the image.

| Fold | Melanoma Jaccard Index | Seborrheic Keratosis Jaccard Index |
|---|---|---|
| 0 | 0.78931 | 0.91302 |
| 1 | 0.76027 | 0.88483 |
| 2 | 0.73652 | 0.88527 |
| 3 | 0.74524 | 0.90066 |
| 4 | 0.78929 | 0.89517 |
| 5 | 0.69436 | 0.91287 |
| 6 | 0.71388 | 0.91264 |
| 7 | 0.82449 | 0.85908 |
| 8 | 0.75025 | 0.85149 |
| 9 | 0.77327 | 0.88322 |
|   | 0.757688 | 0.889825 |

**Table 2 Cross validation average Jaccard Index by fold**

### Comments
From reviewing the test images for the challenge, it appears that the bright light leak might be present in this set of images. If this is true, it would be interesting to compute the final competition metric without these examples.

If the bright spot pattern is indeed a data leak that exists in the test set of images, I expect the following images to be all of class Seborrheic Keratosis: ISIC_0014567.jpg, ISIC_0014574.jpg, ISIC_0014575.jpg, ISIC_0014587.jpg, ISIC_0014588.jpg, ISIC_0014590.jpg, ISIC_0014600.jpg, ISIC_0014619.jpg, ISIC_0014626.jpg, ISIC_0014627.jpg, ISIC_0014629.jpg, ISIC_0014631.jpg, ISIC_0014634.jpg, ISIC_0014643.jpg, ISIC_0014647.jpg, ISIC_0014648.jpg, ISIC_0014649.jpg, ISIC_0014653.jpg